\documentclass[10pt]{article} 
\usepackage{arxiv}

\usepackage{graphicx}
\usepackage{cite}
\usepackage{amsmath,amssymb,amsfonts}
\usepackage{algorithmic}
\usepackage{graphicx}
\usepackage{textcomp}
\usepackage[table]{xcolor}
\usepackage{multirow}
\usepackage{array}
\usepackage{tabularray}
\usepackage{hhline}
\usepackage{hyperref}
\usepackage{url}

\hypersetup{ colorlinks=true, linkcolor=black, filecolor=black, urlcolor=black }

\title{Combining Reinforcement Learning and Behavior Trees for NPCs in Video Games with AMD Schola}

\author{
  Tian Liu, Alex Cann, Ian Colbert, Mehdi Saeedi \\
  Advanced Micro Devices (AMD)\\
  \texttt{\{tianyliu, alexcann, ian.colbert, mehdi.saeedi\}@amd.com} \\
}

\begin{document}

\maketitle

\begin{abstract}
\setcounter{footnote}{0}
While the rapid advancements in the reinforcement learning (RL) research community have been remarkable, the adoption in commercial video games remains slow. In this paper, we outline common challenges the Game AI community faces when using RL-driven NPCs in practice, and highlight the intersection of RL with traditional behavior trees (BTs) as a crucial juncture to be explored further.
Although the BT+RL intersection has been suggested in several research papers, its adoption is rare.
We demonstrate the viability of this approach using AMD Schola---a plugin for training RL agents in Unreal Engine---by creating multi-task NPCs in a complex 3D environment inspired by the commercial video game ``The Last of Us". We provide detailed methodologies for jointly training RL models with BTs while showcasing various skills.
\end{abstract}


\large

\section{Introduction}
Despite the progress in reinforcement learning (RL), the development of advanced non-player characters (NPCs) capable of performing multiple complex tasks remains a significant challenge in practical video game design. 
For example, a recent study \cite{aytemiz2021acting} concludes that NPCs based on behavior trees (BTs) are still more viable than those based on machine learning (ML), calling for new approaches, strategies, and tooling to overcome the barrier to adoption. Additional work has also underscored the need for reusable and adjustable models \cite{jacob2020itsunwieldytakeslot}, motivated by game developers' preferences to reuse previously developed assets, provided that reuse does not result in repetitive gameplay. 

Traditional BT approaches and modern RL techniques each have their respective strengths and limitations in video game development. BTs offer a structured and hierarchical method for managing NPC behaviors, enabling the design of complex systems with predictable outcomes given sufficient development time. However, this complexity can make multi-task BTs less engaging and cumbersome to develop  \cite{jacob2020itsunwieldytakeslot}. 
Conversely, RL provides a dynamic and adaptive approach to decision making \cite{pearce2022counter}, allowing developers to guide an agent through trial-and-error. However, training generally-capable RL models remains a challenge, particularly due to reward shaping, negative task transfer\cite{deramo2024sharingknowledgemultitaskdeep, sodhani2021multi}, and compute resource demands \cite{DBLP:journals/corr/abs-1909-07528}.

To address the complexity of designing RL-based NPCs, researchers have explored adding more parameters to models for training \cite{mclean2025multitaskreinforcementlearningenables} or leveraging large foundation models \cite{reed2022generalistagent}, which are both known to significantly enhance and extend NPC capabilities. However, the increased size and complexity of these models often comes at the cost of increased training duration and high latency during gameplay. 
In addition to the technical challenges that arise when training high-quality RL agents, it is crucial for the resulting NPCs to exhibit consistent behavior to maintain game dynamics \cite{DBLP:conf/cig/ZhangHZCLWL24}. Inconsistent NPC behavior can disrupt the gaming experience for players, leading to frustration and a lack of engagement. Moreover, it adds extra complexities for developers during game quality testing, as they need to ensure that NPCs behave predictably across various scenarios. Thus, consistency and human-like behavior in NPCs \cite{10.1109/CoG51982.2022.9893600, milani2023navigates, campa2025path} are crucial for maintaining game quality and enhancing user experience.

While the benefits of using RL models in video games are clear, the path towards practical use is not straightforward\footnote{There are also non-technical aspects (e.g., data/model ownership) that are out of scope of this evaluation.}. In fact, the gaming industry remains cautious about the adoption of AI in general despite the numerous advancements across the field \cite{futureOfGameDevAI}, suggesting other potential limitations such as insufficient tooling that lacks interpretability and control. 
In this paper, we highlight a BT+RL hybrid NPC as a viable approach to capturing the enhanced abilities of RL with the interpretability of BTs, while reducing the repetitiveness of BTs.
In contrast to existing work, which focuses on research scenarios or safety-critical situations \cite{Li2024, DBLP:journals/corr/abs-1809-10283, DBLP:conf/cig/ZhangHZCLWL24}, we focus on demonstrating the benefits of this approach to game developers. To do so, we design an agent and environment to replicate specific skills inspired by the Human Enemy AI in ``The Last of Us" video game \cite{mcintosh2015human}. We use Unreal Engine and the open-source AMD Schola plugin \cite{schola} to highlight potential methods for combining BT and RL by building  on prior work\footnote{We exclude works that utilize RL for enhancing BT design as this topic is orthogonal from the perspective of enabling developers to utilize RL for controlling NPC actions.}\cite{BTRL2015, fu2016reinforcement, Li2024} and demonstrate how integrating RL into BTs addresses common challenges with both methods. To encourage community investigation, we open-source our environments, models, and implementations in AMD Schola \cite{schola}.

\section{Multi-Skill BT+RL NPCs} \label{sec:multiskill}
To demonstrate the effectiveness of integrating RL models into BTs, we draw inspiration from the Enemy AI in “The Last of Us” \cite{mcintosh2015human}, a critically acclaimed game by Naughty Dog studios. Our goal is to develop an NPC capable of exhibiting a range of skills, including \textbf{Flee} (the NPC tries to create distance between itself and the player), \textbf{Search} (the NPC searches for a target near a  point of interest), \textbf{Combat} (the NPC aims and shoots at the player), \textbf{Hide} (the NPC attempts to stay out of the player’s line of sight), and \textbf{Move} (the NPC navigates to a specified location). Note that these skills are crucial and commonly found in various video games \cite{pearce2022counter, mcintosh2015human, DBLP:journals/corr/abs-2011-04764, aytemiz2021acting}. 

\begin{figure}[tb]
    \centering
    \includegraphics[width=0.5\linewidth]{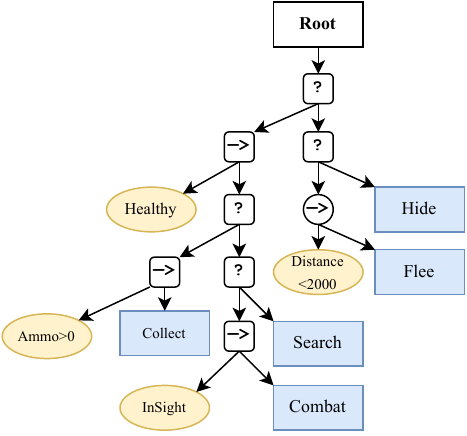}
    \caption{We show the strategic decisions for different skills. Blue receptacles represent skills controlled by RL-based models. ``$?$'' and ``-$>$'' are Selector and Sequence nodes, respectively.} 
    \label{fig:enter-label}
\end{figure}

\begin{figure}[tb]
    \centering
    \includegraphics[width=0.6\linewidth]{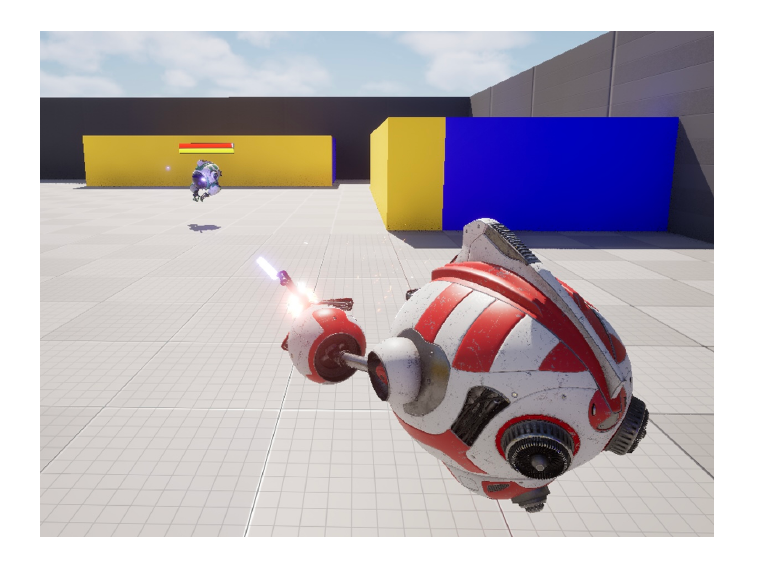}
    \caption{Demonstration of the NPC focusing on the Combat Skill. } 
    \label{fig:npc_visualization}
\end{figure}

Figure~\ref{fig:enter-label} illustrates the BT used for all NPCs in this work. Here, ``Distance'' refers to the distance between the NPC and its opponent. ``InSight'' indicates whether the NPC and the player have a direct line of sight, ``Healthy'' indicates whether an agent has or recently had less than half its health, ``Ammo'' denotes the ammunition count of the NPC, which is required for shooting. Figure \ref{fig:npc_visualization} visualizes the Combat agent. 

\section{RL-based Models} \label{sec:rl_proposed}
For each skill, we use a set of standard observations and actions. Some skills, however, include additional observations and actions. Table \ref{tab:rl_configuration} details the observations, actions, and model architectures.

\newcommand*{\arrg}{\arrayrulecolor{gray!20}}
\newcommand*{\arrb}{\arrayrulecolor{black}}

\begin{figure*}[t!]
    \centering
    \includegraphics[width=\linewidth]{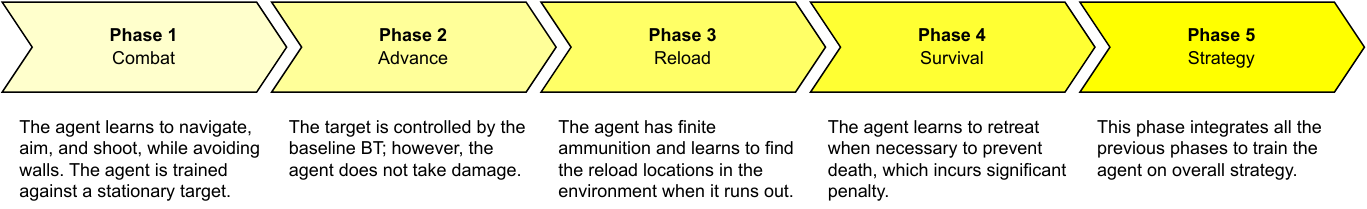}
    \caption{Visualization of the learning curriculum for Curriculum RL agent.}
    \label{fig:curriculum}
\end{figure*}

\begin{table*}[tb]
\centering
    \caption{Rewards and steps used to train different skills. Terminal rewards are denoted by $*$. }. 
    \begin{tabular}{|l|c|l|l|l|c|}
        \hline
        \rowcolor{gray!20}
        & \multicolumn{3}{c|}{\textbf{Reward}} & & \\
        \rowcolor{gray!20}  \textbf{Skill} & \textbf{Step} & \centering \textbf{Wall Collision} &  \multicolumn{1}{|c|}{\textbf{Others}}  & \centering \textbf{Terminal Condition} & \textbf{Steps} \\
        \hline
        \cellcolor{gray!20}Flee & \phantom{-}0.001 & \phantom{-}0 & -1.0 if player distance $<$ 1000*  & Player distance $<$ 1000 & 2M \\
        \hline
        \cellcolor{gray!20}Advance &  -0.001  & -0.01 & \phantom{-}1.0 if player in sight* & Player in sight & 4M \\
        \hline
        \cellcolor{gray!20}Combat &  -0.001 & \phantom{-}0 & \begin{tabular}[l]{@{}l@{}}\phantom{-}0.1 if hits player \\ \phantom{-}1.0 if kills player*\end{tabular} & Player health $\leq$ 0  & 2M \\
        \hline
        \cellcolor{gray!20}Hide &  \phantom{-}0.001 & \phantom{-}0 & -1.0 if player in sight* & Player in sight & 10M\\
        \hline
        \cellcolor{gray!20}Collect & -0.001  & -0.01 & \begin{tabular}[l]{@{}l@{}}\phantom{-}1.0 for successful reload* \\ -0.1 for being hit\end{tabular} & Successful reload & 12M\\
        \hline
    \end{tabular}
    \label{tab:reward}
\end{table*}

\subsection{Training Configuration}
We train each skill using proximal policy optimization with the default settings in RLlib \cite{liang2017rllib} with learning rate $3e^{-4}$ using the steps and rewards given in Table~\ref{tab:reward}, capped at a maximum of 2000 steps per episode, as well as the observations, actions and model architectures given in Table~\ref{tab:rl_configuration}. Terminating rewards in Table~\ref{tab:reward} are indicated by an asterisk (*). The environments used to train each skill are detailed below:
(1) \textbf{Flee:} The Flee training environment is characterized by a randomly spawned player and agent, with the target controlled by a BT approaching the agent at a speed of 300 units per second;
(2) \textbf{Advance:} The environment features a randomly spawned player and agent along with small wall segments, and the target remains stationary throughout each episode;
(3) \textbf{Combat:} The environment for training the combat skill features a randomly spawned player and agent, both can rotate but are stationary in position; 
(4) \textbf{Hide:} The environment is characterized by a randomly spawned player, agent, and obstacles to hide behind, and the player is controlled by a BT approaching the agent at a speed of 100 units per second; and 
(5) \textbf{Collect:} The environment features the agent, a player, controlled by a BT, which pursues the agent, a goal location the NPC tries to navigate to, and small wall segments, all randomly spawned. 

\section{Empirical Evaluation}
\subsection{The Environment}
Our evaluation environment is a competitive third-person shooter game created in Unreal Engine. The game consists of two NPCs competing against each other to reduce the opposing NPC's health to 0. Agents can damage each other by shooting projectiles that are unaffected by gravity. The map is a 4000 units\textsuperscript{2} enclosed square containing static obstacles and ammunition reloads. Ammo is placed at 8 points around the map. All NPCs start the game with 100 health points (HP), 10 ammunition, deal 10 HP of damage per attack, have a 0.15 second firing interval, and 600 units per second movement speed. We restart episodes that take more than 10,000 steps; this happens roughly 10\% of the time. 

\subsection{BT Baseline Model}
As a baseline, we implement a pure BT model. This model uses the same tree structure as our BT+RL outlined in Section \ref{sec:multiskill}, but its leaf nodes for task executions are replaced with pre-defined BT tasks designed to mimic the behavior of their RL model counterparts. For example, the Combat task rotates the NPC towards its target and initiates firing, while the Search task moves the NPC towards its target. For the Flee and Hide skills, we utilize the Unreal Engine's environment query system (EQS) to identify the best direction to flee and where to hide. We align the EQS criteria as closely as possible with the reward function of the corresponding task's RL model.

\subsection{Curriculum Learning Baseline}
For comparison with RL methods, we additionally implement an RL model trained to play the game using curriculum learning as a baseline. The observation space and action space for the model is a superset of the individual skills' observation and action spaces. The curriculum consists of a series of environments where each environment targets a specific subset of skills to learn as detailed in Fig.~\ref{fig:curriculum}. Additionally, we attempted to train RL models without a curriculum however we found that they achieved negligible performance, even when we use a model architecture with significantly increased neurons in the MLP layers. Table \ref{table:rlcurriculum} reports the rewards and the total steps per phase for the curriculum agent.

\begin{table*}[tb]

\newcommand{\CoreObservations}{\multirow{-12}{*}{\begin{tabular}{@{}p{3cm}@{}} 36 rays detecting target, obstacles, and ammo reload locations. \\ Floating point observations for current health points, ammunition count, and normalized direction to target. \end{tabular}}}

\newcommand{\CoreActions}{\multirow{-12}{*}{\begin{tabular}{@{}p{2.25cm}@{}} Lateral Movement \\ Forward Movement \end{tabular}}}

\newcommand{\MLP}{\multirow{-8}{*}{\begin{tabular}{@{}p{1.2cm}@{}} Depth 2 \\ Width 128 \end{tabular}}}
\newcommand{\Attention}{\multirow{-10}{*}{\begin{tabular}{@{}p{2.2cm}@{}} Attention Layer with attention dimension 60 \\ and max sequence length 20, attending over last 20 observations\end{tabular}}}

\newcommand{\HideObservations}{\begin{tabular}{@{}p{3cm}@{}} Player can See Agent. \\ Normalized distance to the first object in the direction of the player. \end{tabular}}

\centering
\caption{RL  Configurations}
\scalebox{0.92}{
\begin{tabular}{|c|p{3cm}|p{3cm}|p{1.2cm}|c|c|c|}
\hline \rowcolor{gray!20} & \multicolumn{2}{c|}{\textbf{Observations}} & \multicolumn{2}{c|}{\textbf{Network}} & \multicolumn{2}{c|}{\textbf{Actions}} \\ \hhline{|~|-|-|-|-|-|-|}
\rowcolor{gray!20} \multirow{-2}{*}{\textbf{Model}} & \centering \textbf{Core} & \centering \textbf{Auxiliary} & \centering \textbf{MLP} & \textbf{Attention Layer} & \textbf{Core} & \textbf{Auxiliary} \\  \hline
\cellcolor{gray!20} Combat &  & \cellcolor{gray!50} & \begin{tabular}{@{}p{1.2cm}@{}} Depth 2 \\ Width 64 \end{tabular} & \cellcolor{gray!50} &  & Shoot \\ \hhline{|-|~|~|-|-|~|-|}
\cellcolor{gray!20} Flee &  & \cellcolor{gray!50} &  &  &  & \cellcolor{gray!50} \\  \hhline{|-|~|~|~|~|~|~|}
\cellcolor{gray!20} Search &  & \cellcolor{gray!50} &  &  &  & \cellcolor{gray!50} \\ \hhline{|-|~|-|~|~|~|~|}
\cellcolor{gray!20} Hide &  & \HideObservations &  &  &  & \cellcolor{gray!50} \\ \hhline{|-|~|-|~|~|~|~|}
\cellcolor{gray!20} Collect &  & Normalized direction and distance to nearest ammo reload location &  &  &  & \cellcolor{gray!50} \\ \hhline{|-|~|-|~|~|~|-|}
\cellcolor{gray!20} Curriculum & \CoreObservations & All of the Above & \MLP & \Attention & \CoreActions & Shoot \\
\hline
\end{tabular}
\label{tab:rl_configuration}
}
\end{table*}

\begin{table}[tb]
\centering
\caption{Rewards and training steps for Curriculum agent.}
\begin{tabular}{|l|c|c|}
\hline
\rowcolor{gray!20}
\textbf{Phase} &  \textbf{Reward} &  \textbf{Training Steps} \\
\hline
\cellcolor{gray!20} & \multirow{4}{*}{\begin{tabular}{@{}p{3cm}@{}} Shot landed: +1.0 \\ Wall Collision: -0.01 \\ Shot taken: -0.1 (Unlimited ammunition) \end{tabular}} &  \multirow{2}{*}{6M} \\
\cellcolor{gray!20}\multirow{-2}{*}{Phase 1 Combat} & & \\ \hhline{|-|~|-|}
\cellcolor{gray!20} & & \multirow{2}{*}{2M} \\ 
\cellcolor{gray!20}\multirow{-2}{*}{Phase 2 Advance} & & \\ \hhline{|-|-|-|}
\cellcolor{gray!20}Phase 3 Move & \begin{tabular}{@{}p{3cm}@{}} Shot landed: +1.0 \\ Wall Collision: -0.01 \\ Shot taken: -0.1 \\ Move when empty: +5.0 \\ Step penalty: -0.01 \end{tabular} &  10M \\
\hline
\cellcolor{gray!20}& \multirow{4}{*}{\begin{tabular}{@{}p{3cm}@{}} Shot landed: +1.0 \\ Wall Collision: -0.01 \\ Death penalty: -10.0 \\ Step penalty: -0.001 \end{tabular}} & \multirow{2}{*}{12M} \\
\cellcolor{gray!20}\multirow{-2}{*}{Phase 4 Survival} & & \\ \hhline{|-|~|-|}
\cellcolor{gray!20} & & \multirow{2}{*}{10M} \\
\cellcolor{gray!20}\multirow{-2}{*}{Phase 5 Strategy} & &  \\
\hline
\end{tabular}
\label{table:rlcurriculum}
\end{table}

\section{Results}
\subsection{Model Quality}

To evaluate model skill we compare each method against two opponents, a single \textit{Static NPC} that does not move nor attack, and an \textit{Aggressive NPC} that is controlled by a simplified version of our baseline BT, which never flees or hides, but is augmented to have distinct offensive advantages (\textit{e.g.}, unlimited ammo).
We evaluate agents based on their win rate, total number of steps elapsed, and additionally, against the \textit{Aggressive NPC} by reporting Damage Dealt. 

\begin{figure}[tb]
    \centering
    \includegraphics[width=0.6\linewidth]{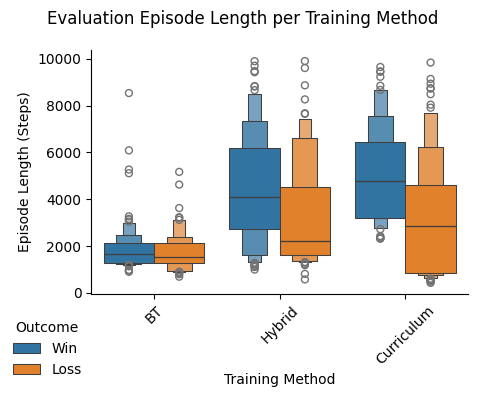}
    \caption{The distribution of episode durations for the wins and losses of each method against the Aggressive NPC.} 
    \label{fig:episode_length_distribution}
\end{figure}

In Table \ref{hthresults}, when comparing success rates, we see that the hybrid approach does significantly better than the curriculum RL model, while performing only slightly worse than the BT model. This result is also reflected in the average damage dealt. The episode durations are plotted in Fig. \ref{fig:episode_length_distribution}, where we see that the BT-based model took both the fewest number of steps in all cases and had similar distributions for both wins and losses. In contrast, both the curriculum model and the hybrid model had much wider distributions of episode length, indicating more variety in episode trajectories.
We note that the hybrid method could benefit from various techniques to enhance RL models, such as curriculum learning or network architectures, as well as improvements to the BT structure.

\subsection{Test time FPS}
To evaluate the test-time performance, we measure the average frames per second (FPS), in the environment and configuration previously used for other experiments, over 100,000 steps. To consider the impact of having multiple model-based NPCs in the scene, we repeat this experiment in an environment with 10 NPCs. As shown in Table \ref{table:inferencefps}, we see that the pure BT approach has the highest average FPS, followed by the curriculum RL model and hybrid model. This follows from the BT utilizing simple computations to compute actions, and the hybrid model computing both a BT and an RL model. In these results, we notably skip model optimizations such as batching, which is enabled by the use of small reusable models in the BT+RL approach, and leave that as future work.

\section{Conclusion \& Future Work}

Our study highlights the intersection of reinforcement learning (RL) and behavior trees (BTs) as a promising direction to integrate reliable and cost-effective deep learning-based agents into commercial video games as NPCs.
With BT+RL, we demonstrate how to develop NPCs capable of interesting behaviors and diverse skills without extensive reward shaping and imitation learning. In addition, the models trained are modular and composable, each targeting simple skills that can be reused in a new BT. As the models are subject to the control of the BT, developers can manually control the behavior of the agent where necessary, or adjust the parameters when the agent invokes RL-driven actions to tune the consistency of the agent. This reusability allows for performance optimizations such as batching, or reduced model sizes for simple or repetitive tasks, which can result in better in-game performance. We open-source our approach to encourage reuse and further development within the community.

\begin{table}[tb]
\centering
\caption{Model evaluation results}
\scalebox{0.95}{
\begin{tabular}{|l|c|c|c|c|c|} 
\hline
\rowcolor{gray!20}
 & \multicolumn{2}{c|}{\textbf{Against Static NPC}}  & \multicolumn{3}{c|}{\textbf{Against Aggressive NPC}}\\ 
\cline{2-6}
 \cellcolor{gray!20}\textbf{Setting} & \cellcolor{gray!20}\textbf{Win Rate} &\cellcolor{gray!20} \textbf{Steps} &  \cellcolor{gray!20}\textbf{Win Rate} & \cellcolor{gray!20}\textbf{Steps} & \cellcolor{gray!20}\textbf{Damage} \\
\hline
BT & 1.00 & 1665.80 & 0.59 & 1839.63 & 170.48 \\ 
\hline
Hybrid & 1.00 & 2441.43 & 0.53 & 3969.22 & 149.86 \\ 
\hline
Curriculum & 1.00 & 3056.50 & 0.41 & 3836.95 & 137.80 \\ 
\hline
\end{tabular}
}
\label{hthresults}
\end{table}

\begin{table}[t]
\centering
\caption{Average FPS over 100,000 steps.}
\begin{tabular}{|l|c|c|} 
\hline
\rowcolor{gray!20}
\textbf{Setting} & \textbf{1 Agent} & \textbf{10 Agents} \\ 
\hline
No Model & 267.73 $\pm$ 3.37\phantom{1} & 188.83 $\pm$ 4.14 \\ 
\hline
BT & 261.90 $\pm$ 10.88 & 155.82 $\pm$ 4.31 \\ 
\hline
Hybrid & 211.90 $\pm$ 4.11\phantom{1} & 109.71 $\pm$ 1.88 \\ 
\hline
Curriculum & 215.80 $\pm$ 9.77\phantom{1} & 116.14 $\pm$ 2.54 \\ 
\hline
\end{tabular}
\label{table:inferencefps}
\end{table}

\bibliographystyle{ieeetr}

\end{document}